# An interpretable and trustworthy AI framework for large-scale longitudinal structure–pain association studies using data from the Osteoarthritis Initiative (OAI)

Jincheng Yu[1], Haoyang Li[2], Yiwen Liu[1,2], Shen Liu[2], Rachel Yuanbao Chen[4,5], C. Kent Kwoh[3], Hongxu Ding[1,4*], Xiaoxiao Sun[1,2*]

[1]Statistics & Data Science GIDP, University of Arizona
[2]Department of Epidemiology and Biostatistics, University of Arizona
[3]College of Medicine Tucson, University of Arizona
[4]R. Kent Coit College of Pharmacy, University of Arizona
[5]University High School
[*]Correspondence should be addressed to: HDing@pharmacy.arizona.edu; xiaosun@arizona.edu

**Author contributions:**
JY: Data curation; Formal analysis; Methodology; Visualization; Writing - original draft
HL: Data curation; Formal analysis; Validation; Visualization; Writing - original draft
YL: Conceptualization; Methodology; Writing - original draft; Writing - review & editing
SL & RYC: Data curation; Writing - original draft
CKK: Conceptualization; Writing - review & editing
HD: Conceptualization; Supervision; Writing - original draft; Writing - review & editing
XS: Conceptualization; Methodology; Supervision; Writing - original draft; Writing - review & editing

***Key Points***

1. An interpretable AI framework combining deep learning–based MRI Osteoarthritis Knee Score prediction with uncertainty quantification increased prediction accuracy, with Matthews correlation coefficients improving from 0.69 to 0.91 for bone marrow lesions, from 0.45 to 0.80 for cartilage loss, and from 0.59 to 0.89 for meniscal extrusion.

2. Applying uncertainty-based filtering enabled expansion of the analytical data to 2,175 knees, supporting robust longitudinal modeling of structure–pain relationships using four complementary pain measures.
3. Bone marrow lesions, cartilage loss, and meniscal extrusion were significantly associated with rapid pain progression, with odds ratios (95% confidence intervals) of 1.62 (1.12–2.35), 1.83 (1.24–2.70), and 2.50 (1.75–3.57), respectively.

***Summary Statement***

This study developed a trustworthy and interpretable AI framework that integrates MRI-based structural prediction with longitudinal statistical modeling to investigate structure–pain relationships in knee osteoarthritis.

# Abstract

***Purpose***: To develop an interpretable and trustworthy AI framework that combines deep learning based MRI Osteoarthritis Knee Score (MOAKS) prediction with interpretable statistical modeling to study structure–pain relationships at scale using data from the Osteoarthritis Initiative (OAI).

***Materials and Methods***: We first developed a deep learning framework to predict MOAKS features directly from knee MRIs and incorporated conformal prediction to provide prediction uncertainty quantification. This uncertainty-aware strategy enables explicit filtering of model outputs, retaining only high-confidence MOAKS predictions at the knee level. Second, we applied a longitudinal latent class mixed model (LCMM) to examine associations between key structural abnormalities and four complementary knee pain measurements.

***Results***: Among the three MRI-defined abnormalities (i.e., bone marrow lesions (BML), cartilage loss (CART), and meniscal extrusion (ME)), our framework substantially improved the Matthews correlation coefficient (MCC) and some other metrics. For example, MCC increased from 0.69 to 0.91 for BML, from 0.45 to 0.80 for CART, and from 0.59 to 0.89 for ME. Using these high-confidence predictions, we expanded the sample size to 2,175 knees for the LCMM analysis. Two distinct pain trajectories were identified (rapid and stable pain progression). The estimated odds ratios (95% CI) for the rapid progression group were 1.62 (1.12–2.35) for BML, 1.83 (1.24–2.70) for CART loss, and 2.50 (1.75–3.57) for ME.

***Conclusion***: These results highlight the importance of these structural abnormalities as risk factors for pain and functional progression in osteoarthritis.

# 1.Introduction

Knee osteoarthritis (OA) is a chronic, degenerative joint disease characterized by the progressive breakdown of articular cartilage, structural changes in subchondral bone, and alterations in periarticular soft tissues, resulting in pain, stiffness, and reduced mobility [1], [2], [3], [4]. As a multifactorial disease, the development and progression of OA are influenced by a wide range of structural abnormalities. As a non-invisive imaging technique, magnetic resonance imaging (MRI) enables sensitive characterization of these abnormalities such as bone marrow lesions (BML), cartilage loss (CART), and meniscal extrusion (ME) [5], [6], [7], [8]. While MRI-defined structural features provide detailed characterization of joint pathology, their associations with patient-reported pain are often heterogeneous and incompletely understood [9]. Elucidating these structure–pain relationships is critical for improving phenotypic stratification, identifying clinically meaningful imaging biomarkers, and guiding the development and evaluation of targeted interventions. However, existing MRI-based studies are often constrained by limited sample sizes, rely on cross-sectional analyses, and frequently use a single pain outcome, limiting their ability to capture the heterogeneity and longitudinal dynamics of structure–pain relationships [10], [11], [12], [13], [14].

In addition, although advanced deep learning models for medical images demonstrate strong predictive performance [2], [10], [15], the uncertainty associated with their predictions has not been adequately evaluated. Owing to their black-box nature and/or independence on distributional assumptions, most existing approaches provide limited insight into prediction reliability [16], [17]. This limitation is particularly critical in multi-stage studies, such as downstream association analyses based on the predictions, where uncertainty in predicted labels may propagate and introduce unknown bias in subsequent analyses [18]. To address these challenges, we propose a two-stage, interpretable, and trustworthy framework to investigate longitudinal structure–pain associations in knee OA, see Figure 1.

In the first stage, we develop a transformer-based deep learning model [19], [20] integrated with conformal prediction to generate MRI-derived structural labels with valid uncertainty measurements [21]. This approach enables distribution-free uncertainty quantification and allows explicit selection of high-confidence predictions for downstream analyses, thereby enhancing the reliability of subsequent inference. In the second stage, we apply longitudinal latent class mixed models (LCMM) to evaluate associations between existing expert-reader and high-confidence predicted labels of structural abnormalities and pain trajectories [22], [23]. This modeling strategy provides interpretable estimates of structure–pain relationships while accounting for population heterogeneity and distinct longitudinal pain progression patterns. Importantly, we incorporate four complementary knee symptom measurements that capture different dimensions of pain severity and functional impact [24], [25], [26]. Modeling these measurements jointly improves robustness, enhances clinical interpretability, and provides a more comprehensive characterization of structure–pain associations.

In summary, the proposed framework enables scalable, trustworthy, and interpretable investigation of longitudinal structure–pain relationships in knee OA. By addressing key limitations of prior studies, such as limited uncertainty quantification, cross-sectional designs, and reliance on single pain measures, it establishes a rigorous foundation for studying the etiology of knee pain in OA.

# 2.Materials and Methods

In this retrospective study, all data used is from the Osteoarthritis Initiative (OAI) [4]. We would like to summarize our framework as a two stage implementation. First, we employed a semi-supervised framework to maximize the utility of the unlabeled OAI MRIs and generated precise predictions for these data [19]. Second, we applied a cross-conformal prediction method to quantify the prediction uncertainty [21], retaining highly confident predictions. Using these high-confidence predictions and MRI readings from radiologists, we then implemented association study using the latent class mixed models (LCMM) [23]. The data flow for the two stages are shown in Figure 2 and Figure 3. In the following, we briefly describe the proposed framework. For additional details, please refer to the supplementary materials.

## 2.1.Semi-supervised Framework

### 2.1.1.Self-supervised Pretraining

Our masked autoencoder (MAE) architecture was composed of 12 stacked vision transformer (ViT) layers [27]. A standard ViT block typically consisted of an attention layer, multiple linear transformation layers, and a feedforward layer. With a 75% mask ratio setting during training, MAE could achieve efficient and robust representation learning.  Unlike the original MAE framework, which includes a *[CLS]* token for classification, we omitted this token, as our focus was exclusively on slice-level representation learning.

### 2.1.2.Supervised Training

Prediction of knee abnormalities cannot rely on individual MRI slices, as these pathologies often extend across the full three-dimensional structure. Therefore, we employed a standard 3D ResNet architecture to integrate information across all slices, enabling joint representation learning and generating a subject-level prediction for the entire knee. Since the data in the OAI are imbalanced, we adopted a hybrid strategy combining focal loss and mini-batch oversampling to mitigate this issue [28], [29]. In particular, the focal loss is defined as:

$$FL(p_t) = -(1-p_t)^{\gamma} log(p_t),$$

where $p_t$ denotes the predicted probability of the ground-truth class, and $\gamma$ is a weighting coefficient with a default value of 2.5 [29]. Besides, we used mini-batch oversampling to keep a consistent ratio of negative to positive samples in every batch. The details to determine best ratio are shown in the supplemental materials.

### 2.1.3.Data Preprocessing

**Self-supervised Learning.** We extracted all available 2D slices from the sagittal view, yielding approximately 7.5 million slices. We applied the transformation and normalization protocols proposed by MedSAM [30].

**Supervised Learning.** Our experimental setup focuses on two key components: label definition and dataset construction. For labels, we used the MOAKS scoring system (0-3 scale) and, following previous studies [10], [15], simplified the scores into binary classes: 0 is negative, and 1-3 are positive. The MOAKS system provides a rigorous and standardized framework for MRI assessment. Following a structured training and calibration session, two expert readers independently evaluated MRI scans from 20 knees. To assess intra-reader reliability, one reader re-evaluated the same 20 scans after a 4-week interval, with the images presented in randomized order. For the dataset construction, we used the 10-fold cross-validation. To prevent data leakage from participants who had multiple scans over time, we split the data by patient IDs rather than by individual knees, ensuring no participants appeared in both the training and testing datasets.

### 2.1.4.Evaluation

In addition to the area under the ROC curve (AUC), we evaluated model performance using the F1 score, balanced accuracy (BAcc), and Matthews correlation coefficient (MCC), all of which are well-suited for assessing performance in imbalanced classification settings. We denote TP, TN, FP and FN as true positive, true negative, false positive and false negative. The equations for these matrices are listed below:

$$F1\ score = \frac{2TP}{2TP + FP + FN},$$

$$BAcc = \frac{1}{2}\left(\frac{TP}{TP + FN} + \frac{TN}{TN + FP}\right),$$

$$MCC = \frac{TP \cdot TN - FP \cdot FN}{\sqrt{(TP+FP)(TP+FN)(TN+FP)(TN+FN)}}.$$

## 2.2.Trustworthy Gate

We leveraged the conformal prediction method as a trustworthy gate for retaining highly confident predictions. The conformal prediction method evaluates both potential labels, 0 and 1, for each test data point. For each postulated label, it tests the hypothesis of exchangeability by measuring how well the test example conforms to the calibration set [21]. The procedure outputs the corresponding p-values, $p^0$ and $p^1$. Given a significance level $\alpha$, the prediction set for a test sample is constructed as $\{y \mid y \in \{0, 1\}: p^y > \alpha\}$. The sample with a single label in the prediction set would be recognized as the high confidence sample [21]. Within a K-fold cross validation framework, we utilized the cross-conformal prediction (CCP) to ensemble the results from the K models, providing a more robust and informative output. Specifically, we calculated the cross-conformal p-value for the postulated label y as:

$$p^y := \frac{\sum_{k=1}^{K} |\{i \in I_k : \alpha_{i,k} \le \alpha_k^y\}| + 1}{|I| + 1},$$

For the k-th fold, the training split was used to train the k-th model, while the complementary calibration dataset, denoted as $I_k$, was used to calculate the conformal scores $\alpha_{\cdot,k}$. $|I|$ denotes the size of the entire labeled set. In this study, we leveraged the scoring criteria from [21]:

$$a_{i,k}(p, y) = y \cdot p + (1 - y) \cdot (1 - p),$$

where $p$ is the probability predicted by the 3D ResNet model and $y$ is the ground-truth label of the data point from the calibration set.

## 2.3.Association Analysis

### 2.3.1.Risk Factors Associated with Pain Phenotypes

The primary risk factors in this study were the MRI-defined abnormalities (i.e., BML, CART and ME) [31]. In addition, body mass index (BMI), age, sex, race, history of knee injury, history of knee surgery, mean follow-up physical activity scale for the elderly (PASE), comorbidity (COMORB), and center for epidemiologic studies depression scale (CESD) scores were also included as risk factors in our association model.

### 2.3.2.Latent Class Mixed Model (LCMM)

Latent Class Mixed Model (LCMM) is a statistical framework based on linear mixed models, capable of identifying latent groups of samples exhibiting similar trajectories according to selection criteria, and simultaneously assessing the association between them with preselected risk factors [22]. Specifically in our study, we leveraged four knee symptom measurements as our selection criteria, see Table 1. With the preselected risk factors incorporated in this model, we could estimate the probability of each knee belonging to each latent group and assign the group by the highest posterior probability [22]. The data flow for the association analysis is detailed in Figure 3.

# 3.Results

## 3.1.Prediction Performance

To evaluate the performance of our semi-supervised learning framework and the CCP method. We compared our model ('Ours' in Table 2) with the existing models [10], [20]. From Table 2, our framework showed superior performance across all the abnormalities. Setting the significance level $\alpha$ at 0.1, the predictive performance on these high-confidence subsets was evaluated using the metrics detailed in Section 3.1.4. From Figure 4, the predictive performance improved significantly with the trustworthy gate.

## 3.2.Trajectory Patterns Identified by LCMM

Across models for BML, CART, and ME, two latent classes were identified for each model. Over 9 years, a consistent trajectory pattern was observed across all three models, shown in Figure 5. One trajectory is characterized by persistently low levels of four knee symptom measurements, indicating a stable symptom course. The other trajectory shows progressive worsening across all four measurements over time.

### 3.3. Baseline Characteristics by Trajectory Patterns

Baseline characteristics differed between the two latent trajectory classes (Table 3). Compared with the stable pain trajectory class, the rapid progression trajectory class had a higher proportion of Black participants (16% vs. 8.1%, $p < 0.001$) and a higher BMI distribution, with a greater proportion classified as obese (44% vs. 28%) and a smaller proportion in the normal-weight range (12% vs. 32%; $p < 0.001$). The age distribution also differed between the two trajectory classes, with a larger proportion of participants aged >65 years in the rapid progression trajectory class (41% vs. 34%; $p = 0.047$). By contrast, sex, prior knee injury, and prior knee surgery did not differ significantly between the two trajectory classes.

Baseline clinical characteristics also differed between the two trajectory classes, with the rapid pain progression class showing lower mean PASE scores (144.2 vs. 159.4, $p < 0.001$), higher mean comorbidity burden (0.7 vs. 0.4, $p < 0.001$), and higher mean CESD scores (9.5 vs. 5.5, $p < 0.001$). Structural imaging features were also more common in the rapid pain progression trajectory class: baseline BML was present in 67% versus 50%, cartilage loss in 73% versus 55%, and meniscal extrusion in 56% versus 31% (all p-values < 0.001). Overall, the two trajectory classes differed with respect to baseline structural abnormalities and clinical characteristics, with the rapid pain progression class showing a more adverse profile.

### 3.4.Association Analysis between Risk Factors and Trajectories

The results of the class membership model within the LCMM framework are summarized in Table 4. Across all three models, higher baseline BMI (OR=1.10, 95% CI: 1.06-1.14) and higher mean CESD (OR=1.10, 95% CI: 1.07-1.13) were consistently associated with greater odds of belonging to the rapid pain progression trajectory. All three MRI-defined structural features were also significantly associated with the rapid pain progression trajectory when examined separately; the existence of these features indicates the higher odds of belonging to the rapid progression trajectory. Among the three imaging features, ME showed the strongest association (OR=2.50, 95% CI: 1.75-3.57). Black race was additionally associated with higher odds of the rapid pain progression trajectory in the ME model (OR = 1.64, 95% CI: 1.01-2.67).

## 4.Discussion

Structural abnormalities assessed by MRI, such as bone marrow lesions and cartilage loss, play a critical role in knee osteoarthritis [32], [33], however, their associations with knee pain remain inconsistent and difficult to characterize due to population and symptom heterogeneity and limited sample sizes. This study presents an interpretable and trustworthy AI framework that integrates uncertainty-aware deep learning with longitudinal latent class modeling to advance the investigation of structure–pain relationships in knee osteoarthritis. Importantly, MRI-defined abnormalities, particularly meniscal extrusion, were strongly associated with the rapid pain progression trajectory, supporting their role as key imaging biomarkers. In addition, non-imaging factors such as BMI and depressive symptoms were associated with worse outcomes, emphasizing the multifactorial nature of pain progression in osteoarthritis. Together, these findings demonstrate that combining scalable AI-derived imaging features with interpretable statistical models can enhance our understanding of disease heterogeneity and provide a foundation for more precise risk stratification and targeted intervention in osteoarthritis. In this study, we focused on three imaging features due to the limited availability of labeled MRI data for other MRI-defined features (e.g., effusion-synovitis). In future work, we plan to incorporate additional training data from large-scale studies, such as Multicenter Osteoarthritis Study (MOST), to expand the analysis to a broader set of imaging features.

# Reference

[1] S. Tang *et al.*, “Osteoarthritis,” *Nat. Rev. Dis. Primer*, vol. 11, no. 1, p. 10, Feb. 2025, doi: 10.1038/s41572-025-00594-6.

[2] D. Hayashi, F. W. Roemer, and A. Guermazi, “Osteoarthritis year in review 2024: Imaging,” *Osteoarthritis Cartilage*, vol. 33, no. 1, pp. 88–93, Jan. 2025, doi: 10.1016/j.joca.2024.10.009.

[3] H. Harandi, S. Mohammadi, and A. Guermazi, “The 18th international workshop on osteoarthritis imaging. Structure and pain: Where are we today?,” *Osteoarthr. Imaging*, vol. 5, no. 3, p. 100356, Sep. 2025, doi: 10.1016/j.ostima.2025.100356.

[4] J. B. Driban *et al.*, “The state of the Osteoarthritis Initiative (OAI): Entering a new era,” *Semin. Arthritis Rheum.*, vol. 75, p. 152887, Dec. 2025, doi: 10.1016/j.semarthrit.2025.152887.

[5] D. Herrera, A. Almhdie-Imjabbar, H. Toumi, and E. Lespessailles, “Magnetic resonance imaging-based biomarkers for knee osteoarthritis outcomes: A narrative review of prediction but not association studies,” *Eur. J. Radiol.*, vol. 181, p. 111731, Dec. 2024, doi: 10.1016/j.ejrad.2024.111731.

[6] M. S. Pedoia V Norman B, Mehany SN, Bucknor MD, Link TM, “3D convolutional neural networks for detection and severity staging of meniscus and PFJ cartilage morphological degenerative changes in osteoarthritis and anterior cruciate ligament subjects,” *J Magn Reson Imaging*, vol. 49, no. 2, pp. 400–410, 2019.

[7] Z. X. Hu J Zheng C, Yu Q, Zhong L, Yu K, Chen Y, Wang Z, Zhang B, Dou Q., “DeepKOA: a deep-learning model for predicting progression in knee osteoarthritis using multimodal magnetic resonance images from the osteoarthritis initiative.,” *Quant Imaging Med Surg*, vol. 13, no. 8, pp. 4852–4866, 2023.

[8] Z. Qiu *et al.*, “Learning Co-Plane Attention Across MRI Sequences for Diagnosing Twelve Types of Knee Abnormalities,” *Nat. Commun.*, vol. 15, no. 1, p. 7637, 2024.

[9] S. Liu, X. Sun, Y. Ge, T. N. Duong, and C. K. Kwoh, “ASSOCIATION OF MRI-DEFINED STRUCTURE FEATURES AT BASELINE WITH KNEE PAIN TRAJECTORIES,” *Osteoarthr. Imaging*, vol. 4, p. 100187, 2024, doi: 10.1016/j.ostima.2024.100187.

[10] N. Bien *et al.*, “Deep-learning-assisted diagnosis for knee magnetic resonance imaging: Development and retrospective validation of MRNet,” *PLOS Med.*, vol. 15, no. 11, pp. 1–19, 2018.

[11] F. Liu *et al.*, “Deep Learning Approach for Evaluating Knee MR Images: Achieving High Diagnostic Performance for Cartilage Lesion Detection,” *Radiology*, vol. 289, no. 1, pp. 160–169, 2018.

[12] M. S. Raman S. ,. Gold, G. E. ,. Rosen *et al.*, “Automatic estimation of knee effusion from limited MRI data,” *Sci. Rep.*, vol. 12, no. 1, 2022.

[13] K. V. Chang GH Felson DT, Qiu S, Guermazi A, Capellini TD, “Assessment of knee pain from MR imaging using a convolutional Siamese network,” *Eur. Radiol.*, vol. 30, no. 12, 2020.

[14] L. D. Tack A Shestakov A. and Z. S, “A Multi-Task Deep Learning Method for Detection of Meniscal Tears in MRI Data from the Osteoarthritis Initiative Database.,” *Front Bioeng Biotechnol*, vol. 9, 2021.

[15] S. Liu *et al.*, “Comparison of evaluation metrics of deep learning for imbalanced imaging data in osteoarthritis studies,” *Osteoarthritis Cartilage*, vol. 31, no. 9, pp. 1242–1248, Sep. 2023, doi: 10.1016/j.joca.2023.05.006.

[16] S. Faghani *et al.*, “Quantifying Uncertainty in Deep Learning of Radiologic Images,” *Radiology*, vol. 308, no. 2, p. e222217, Aug. 2023, doi: 10.1148/radiol.222217.

[17] S. Faghani, C. Gamble, and B. J. Erickson, “Uncover This Tech Term: Uncertainty Quantification for Deep Learning,” *Korean J. Radiol.*, vol. 25, no. 4, p. 395, 2024, doi: 10.3348/kjr.2024.0108.

[18] S. Faghani *et al.*, “Mitigating Bias in Radiology Machine Learning: 3. Performance Metrics,” *Radiol. Artif. Intell.*, vol. 4, no. 5, p. e220061, Sep. 2022, doi: 10.1148/ryai.220061.

[19] K. He, X. Chen, S. Xie, Y. Li, P. Dollár, and R. Girshick, “Masked Autoencoders Are Scalable Vision Learners,” Dec. 19, 2021, *arXiv*: arXiv:2111.06377. doi: 10.48550/arXiv.2111.06377.

[20] C. Feichtenhofer, H. Fan, J. Malik, and K. He, “SlowFast Networks for Video Recognition,” Oct. 29, 2019, *arXiv*: arXiv:1812.03982. doi: 10.48550/arXiv.1812.03982.

[21] V. Vovk, “Cross-conformal predictors,” *Ann. Math. Artif. Intell.*, vol. 74, no. 1–2, pp. 9–28, Jun. 2015, doi: 10.1007/s10472-013-9368-4.

[22] S. Liu, X. Sun, Y. Ge, T. N. Duong, and C. K. Kwoh, “Association of baseline MRI-defined structural features with knee symptom trajectories over nine years: Data from the osteoarthritis initiative (OAI),” *Osteoarthr. Imaging*, p. 100380, Nov. 2025, doi: 10.1016/j.ostima.2025.100380.

[23] C. Proust-Lima, V. Philipps, and B. Liquet, “Estimation of Extended Mixed Models Using Latent Classes and Latent Processes: The *R* Package **lcmm**,” *J. Stat. Softw.*, vol. 78, no. 2, 2017, doi: 10.18637/jss.v078.i02.

[24] E. M. Roos and L. S. Lohmander, “The Knee injury and Osteoarthritis Outcome Score (KOOS): from joint injury to osteoarthritis”.

[25] S. M. Nugent, T. I. Lovejoy, S. Shull, S. K. Dobscha, and B. J. Morasco, “Associations of Pain Numeric Rating Scale Scores Collected during Usual Care with Research Administered Patient

Reported Pain Outcomes," *Pain Med.*, vol. 22, no. 10, pp. 2235–2241, Oct. 2021, doi: 10.1093/pm/pnab110.

[26] J. M. Quintana *et al.*, "Health-Related Quality of Life and Appropriateness of Knee or Hip Joint Replacement," *Arch. Intern. Med.*, vol. 166, no. 2, p. 220, Jan. 2006, doi: 10.1001/archinte.166.2.220.

[27] A. Dosovitskiy *et al.*, "An Image is Worth 16x16 Words: Transformers for Image Recognition at Scale," Jun. 03, 2021, *arXiv*: arXiv:2010.11929. doi: 10.48550/arXiv.2010.11929.

[28] R. Shwartz-Ziv, M. Goldblum, Y. L. Li, C. B. Bruss, and A. G. Wilson, "Simplifying Neural Network Training Under Class Imbalance," Dec. 05, 2023, *arXiv*: arXiv:2312.02517. doi: 10.48550/arXiv.2312.02517.

[29] T.-Y. Lin, P. Goyal, R. Girshick, K. He, and P. Dollár, "Focal Loss for Dense Object Detection," Feb. 07, 2018, *arXiv*: arXiv:1708.02002. doi: 10.48550/arXiv.1708.02002.

[30] J. Ma, Y. He, F. Li, L. Han, C. You, and B. Wang, "Segment Anything in Medical Images," *Nat. Commun.*, vol. 15, no. 1, p. 654, 2024.

[31] D. J. Hunter *et al.*, "Evolution of semi-quantitative whole joint assessment of knee OA: MOAKS (MRI Osteoarthritis Knee Score)," *Osteoarthritis Cartilage*, vol. 19, no. 8, pp. 990–1002, Aug. 2011, doi: 10.1016/j.joca.2011.05.004.

[32] S. J. J. Drummen, D. Aitken, P. Otahal, G. Jones, S. M. Bierma-Zeinstra, and J. Runhaar, "The associations between a diagnosis of early-stage knee osteoarthritis and the presence of MRI abnormalities in knees without radiographic osteoarthritis: data from 1,513 knees of the osteoarthritis initiative," *Osteoarthritis Cartilage*, p. S1063458426006849, Feb. 2026, doi: 10.1016/j.joca.2026.02.015.

[33] J. Fu *et al.*, "An MRI-based radiomics framework for early identification and progression stratification in knee osteoarthritis: data from the osteoarthritis initiative," *BMC Musculoskelet. Disord.*, vol. 26, no. 1, p. 1018, Oct. 2025, doi: 10.1186/s12891-025-09234-2.

# Tables

| **Pain Measure** | **Range** |
|---|---|
| **KOOS:** Knee injury and Osteoarthritis Outcome Score for pain (100 = no pain) | 0–100 |
| **WOMAC pain:** Western Ontario and McMaster Universities Osteoarthritis Index – pain subscale (20 = most severe pain) | 0–20 |
| **WOMAC function:** Western Ontario and McMaster Universities Osteoarthritis Index – physical function subscale (68 = worst function) | 0–68 |
| **NRS:** Numeric Rating Scale for pain (10 = most severe pain) | 0–10 |

**Table 1.** Four pain/function measurements used in the association analysis.

| | **BML** | | | | **CART** | | | | **ME** | | | |
|---|---|---|---|---|---|---|---|---|---|---|---|---|
| | AUC | BAcc | F1score | MCC | AUC | BAcc | F1score | MCC | AUC | BAcc | F1score | MCC |
| MRNet | 0.86 (0.03) | 0.66 (0.08) | **0.88** (0.02) | 0.36 (0.11) | | | | | | | | |
| ResNet | 0.87 (0.02) | 0.80 (0.02) | 0.78 (0.02) | 0.61 (0.04) | 0.80 (0.03) | 0.73 (0.03) | 0.88 (0.01) | 0.39 (0.05) | 0.86 (0.02) | 0.78 (0.02) | 0.69 (0.03) | 0.59 (0.04) |
| Ours | **0.91** (0.01) | **0.84** (0.02) | 0.82 (0.02) | **0.69** (0.03) | **0.83** (0.02) | **0.77** (0.02) | **0.89** (0.01) | **0.45** (0.04) | **0.86** (0.01) | **0.79** (0.02) | **0.69** (0.02) | **0.59** (0.02) |
| Ours (CCP) | **0.97** | **0.95** | **0.94** | **0.91** | **0.98** | **0.95** | **0.95** | **0.80** | **0.94** | **0.93** | **0.91** | **0.89** |

**Table 2. Performance comparison.** The table compares the predictive performance of our approach with established deep learning models (MRNet [15] and Resnet [20]). Each cell details the mean and standard deviation of the evaluation metrics.

| **Covariates** | **Overall**[1] | **Stable** N = 1,972 | **Rapid** N = 203 | **p-value**[1] |
|---|---|---|---|---|
| **Gender** | | | | 0.320 |
| Female | 1,214 (56%) | 1,094 (55%) | 120 (59%) | |
| Male | 961 (44%) | 878 (45%) | 83 (41%) | |
| **Race** | | | | <0.001 |
| White | 1,943 (89%) | 1,780 (90%) | 163 (80%) | |
| Black | 192 (8.8%) | 159 (8.1%) | 33 (16%) | |
| Asian | 14 (0.6%) | 13 (0.7%) | 1 (0.5%) | |

| Covariates | Overall[1] | Stable<br>N = 1,972 | Rapid<br>N = 203 | p-value[1] |
|---|---|---|---|---|
| Other | 26 (1.2%) | 20 (1.0%) | 6 (3.0%) | |
| **Age (grouped)** | | | | 0.047 |
| <55 | 688 (32%) | 637 (32%) | 51 (25%) | |
| 55-65 | 737 (34%) | 669 (34%) | 68 (33%) | |
| >65 | 750 (34%) | 666 (34%) | 84 (41%) | |
| **BMI (grouped)** | | | | <0.001 |
| <25 (Normal) | 646 (30%) | 622 (32%) | 24 (12%) | |
| 25-30 (Overweight) | 885 (41%) | 795 (40%) | 90 (44%) | |
| ≥30 (Obese) | 644 (30%) | 555 (28%) | 89 (44%) | |
| **Injury** | | | | 0.609 |
| No | 1,723 (79%) | 1,565 (79%) | 158 (78%) | |
| Yes | 452 (21%) | 407 (21%) | 45 (22%) | |
| **Surgery** | | | | 0.183 |
| No | 2,058 (95%) | 1,870 (95%) | 188 (93%) | |
| Yes | 117 (5.4%) | 102 (5.2%) | 15 (7.4%) | |
| **PASE, Mean (SD)** | 158.0 (64.6) | 159.4 (63.8) | 144.2 (70.1) | <0.001 |
| **Comorbidity, Mean (SD)** | 0.4 (0.8) | 0.4 (0.7) | 0.7 (1.1) | <0.001 |
| **CESD, Mean (SD)** | 5.9 (5.4) | 5.5 (5.0) | 9.5 (7.7) | <0.001 |
| **BML** | | | | <0.001 |
| No | 1,046 (48%) | 980 (50%) | 66 (33%) | |
| Yes | 1,129 (52%) | 992 (50%) | 137 (67%) | |
| **CART** | | | | <0.001 |
| No | 933 (43%) | 879 (45%) | 54 (27%) | |
| Yes | 1,242 (57%) | 1,093 (55%) | 149 (73%) | |
| **ME** | | | | <0.001 |
| No | 1,459 (67%) | 1,370 (69%) | 89 (44%) | |
| Yes | 716 (33%) | 602 (31%) | 114 (56%) | |

[1]p-values were calculated using Pearson's Chi-squared test without continuity correction for Gender, Age group, BMI group, injury, surgery, BML, CART, and ME; Fisher's exact test for Race; and the Wilcoxon rank-sum test for mean PASE, mean comorbidity score, and mean CESD score.

**Table 3. Baseline characteristics by the two latent trajectory classes.** Rapid represents the rapid pain progression trajectory, whereas Stable corresponds to the stable pain trajectory.

| Variable | BML model | | CART model | | ME model | |
|---|---|---|---|---|---|---|
| | OR (95% CI) | p-value | OR (95% CI) | p-value | OR (95% CI) | p-value |
| Baseline BMI (per kg/m^2) | 1.10 (1.06, 1.14) | <0.0001 | 1.10 (1.06, 1.14) | <0.0001 | 1.09 (1.05, 1.13) | <0.0001 |
| BML (Yes vs No) | 1.62 (1.12, 2.35) | 0.0108 | | | | |
| CART (Yes vs No) | | | 1.83 (1.24, 2.70) | 0.0025 | | |
| ME (Yes vs No) | | | | | 2.50 (1.75, 3.57) | <0.0001 |
| Race: Black (vs White) | | | | | 1.64 (1.01, 2.67) | 0.0453 |
| Mean CESD | 1.10 (1.07, 1.13) | <0.0001 | 1.10 (1.07, 1.13) | <0.0001 | 1.09 (1.07, 1.12) | <0.0001 |

**Table 4. LCMM results in association analysis.** Significant predictors only (p <= 0.05). Odds ratios (ORs) are reported for membership in the rapid pain progression trajectory relative to the stable pain trajectory. Each model includes a single imaging feature and adjusts for the same set of covariates.

# Figure Legends

**Figure 1. Overview of the study pipeline.** (A) The semi-supervised framework , once trained, will generate labels for unannotated MRI-defined abnormalities. (B) A "Trustworthy Gate" module employs cross-conformal prediction (CCP) to quantify the uncertainty of initial predictions. (C) The "Trustworthy Gate" will retain the cleaned high-confidence samples for association analysis. (D) A latent class mixed model (LCMM) incorporates both these high-confidence and the originally labeled data to identify significant risk factors contributing to rapid pain progression.

**Figure 2. Data flow for the deep learning stage.**

**Figure 3. Data flow for the association analysis.**

**Figure 4. Performance improvements achieved by incorporating CCP.** The dumbbell plot compares the predictive performance across three structure features. Light blue dots show the introduced model, and red dots show the improved model. Adding CCP consistently improved performance across all metrics for every abnormality.

**Figure 5. Trajectory patterns with preselected measurements.** Panels show KOOS Pain, WOMAC Pain, WOMAC Function, and NRS for the stable and rapid pain progression latent classes in the three models. Solid lines represent posterior probability-weighted mean scores at each follow-up year, and shaded bands denote 95% confidence intervals. To ensure consistent interpretation across measures, all outcomes are harmonized so that higher scores indicate worse symptoms; specifically, KOOS scores are transformed as $100 - \text{KOOS}_{\text{original}}$.

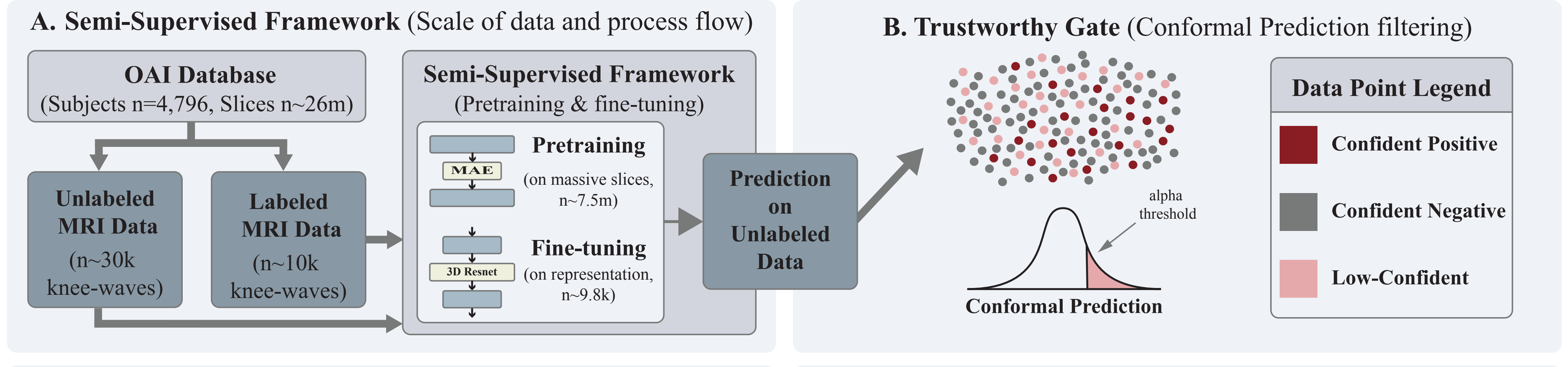
A. Semi-Supervised Framework (Scale of data and process flow)
OAI Database
(Subjects n=4,796, Slices n~26m)
Unlabeled MRI Data
(n~30k knee-waves)
Labeled MRI Data
(n~10k knee-waves)
Semi-Supervised Framework
(Pretraining & fine-tuning)
MAE
Pretraining
(on massive slices, n~7.5m)
3D Resnet
Fine-tuning
(on representation, n~9.8k)
Prediction on Unlabeled Data
B. Trustworthy Gate (Conformal Prediction filtering)
alpha threshold
Conformal Prediction
Data Point Legend
Confident Positive
Confident Negative
Low-Confident

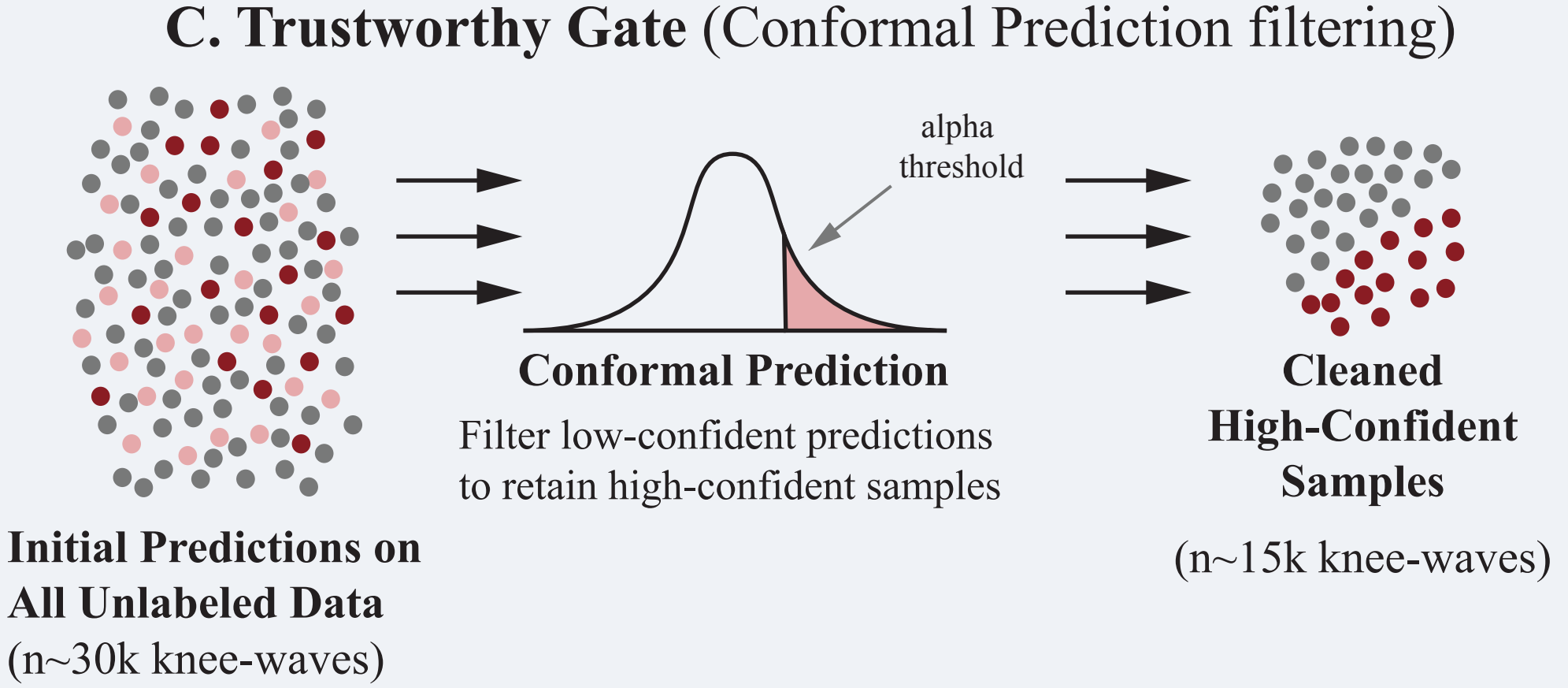
C. Trustworthy Gate (Conformal Prediction filtering)
alpha threshold
Conformal Prediction
Filter low-confident predictions
to retain high-confident samples
Cleaned
High-Confident
Samples
(n~15k knee-waves)
Initial Predictions on
All Unlabeled Data
(n~30k knee-waves)

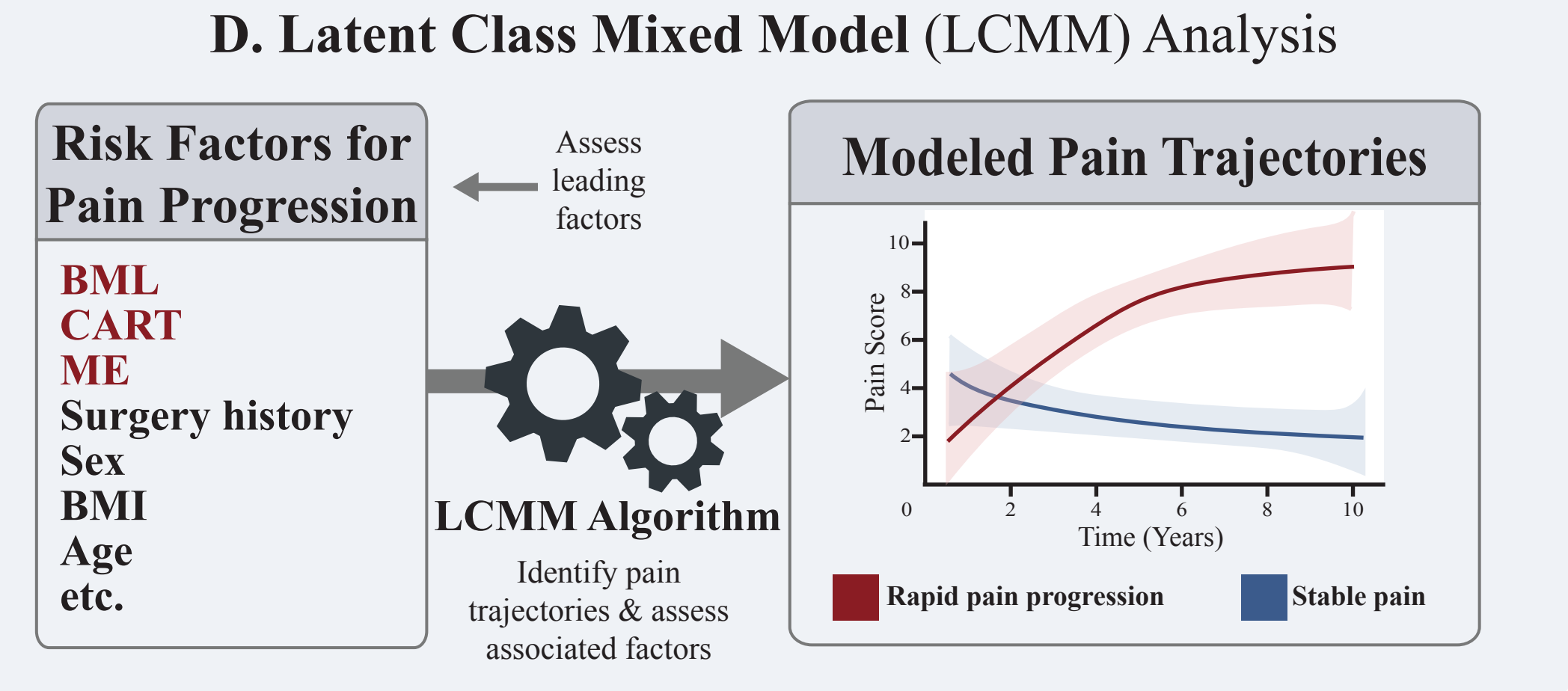
D. Latent Class Mixed Model (LCMM) Analysis
Risk Factors for Pain Progression
BML
CART
ME
Surgery history
Sex
BMI
Age
etc.
Assess leading factors
LCMM Algorithm
Identify pain trajectories & assess associated factors
Modeled Pain Trajectories
Pain Score
Time (Years)
0
2
4
6
8
10
Rapid pain progression
Stable pain

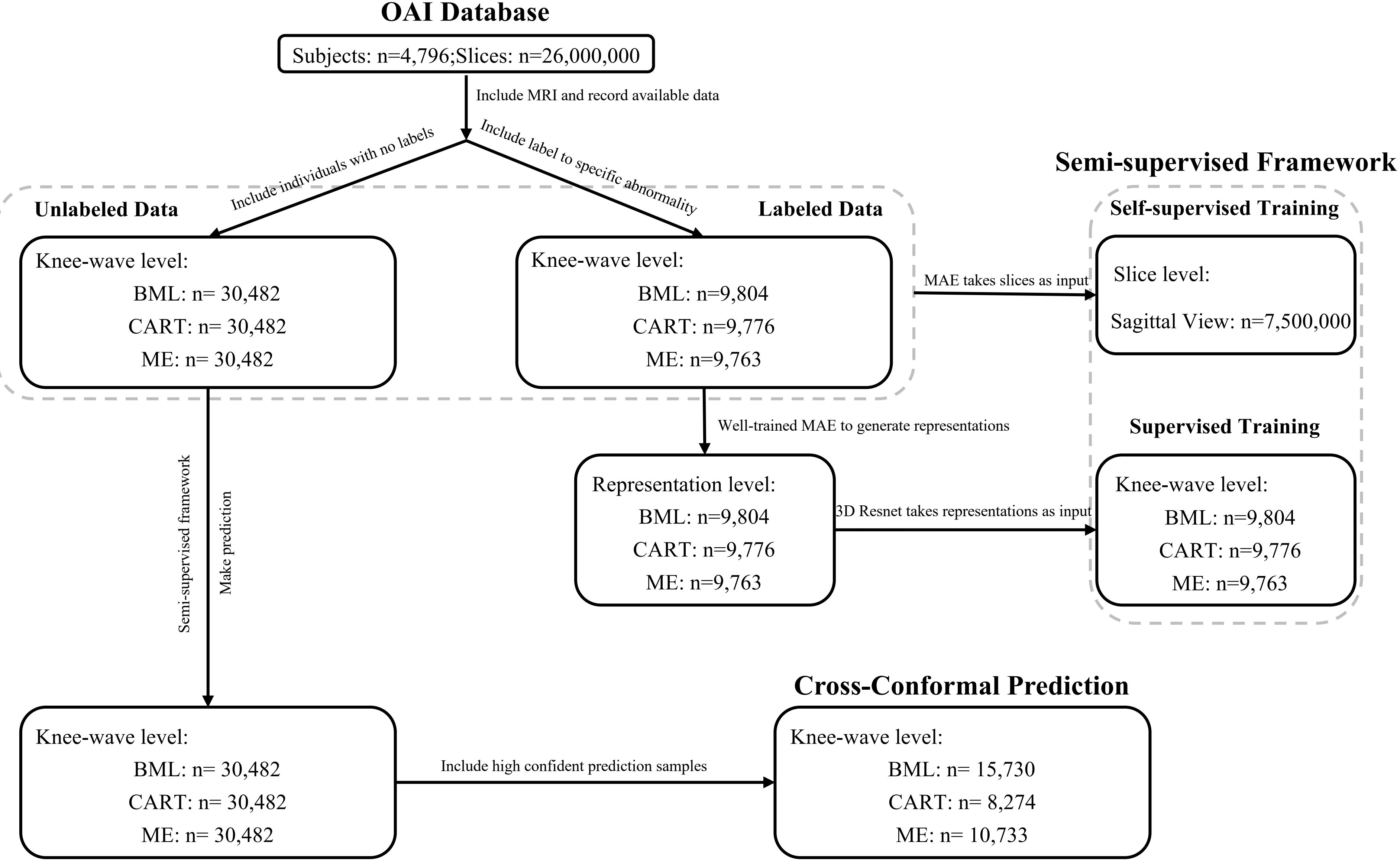

OAI Database
Subjects: n=4,796;Slices: n=26,000,000
Include MRI and record available data
Include individuals with no labels
Include label to specific abnormality
Unlabeled Data
Labeled Data
Knee-wave level:
BML: n= 30,482
CART: n= 30,482
ME: n= 30,482
Knee-wave level:
BML: n=9,804
CART: n=9,776
ME: n=9,763
Semi-supervised Framework
Self-supervised Training
MAE takes slices as input
Slice level:
Sagittal View: n=7,500,000
Well-trained MAE to generate representations
Representation level:
BML: n=9,804
CART: n=9,776
ME: n=9,763
3D Resnet takes representations as input
Supervised Training
Knee-wave level:
BML: n=9,804
CART: n=9,776
ME: n=9,763
Semi-supervised framework
Make prediction
Knee-wave level:
BML: n= 30,482
CART: n= 30,482
ME: n= 30,482
Include high confident prediction samples
Cross-Conformal Prediction
Knee-wave level:
BML: n= 15,730
CART: n= 8,274
ME: n= 10,733

OAI cohort
n = 4,796 participants
9,592 knees

Restricted to knees with at least one baseline symptom measure equal to zero
n = 3,133 participants
4,751 knees
1,515 participants with one knee

Restricted to participants with complete covariate data (e.g., PASE, COMORB, CESD)
n = 2,800 participants
4,207 knees
1,393 participants with one knee

Analytic OAI cohort eligible for imaging linkage
n = 1,733 participants
2,175 knees
1,291 participants with one knee

Prediction-derived imaging data available

MOAKS expert-reading data available

Prediction-derived imaging data available within analytic cohort
n = 1,857

MOAKS expert-reading data available within analytic cohort
n = 1,301 knees

Overlap summarized below

Final linked imaging subsets
Total: n = 2,175
Prediction only: n = 874 knees
MOAKS only: n = 318 knees
Both sources available: n = 983 knees

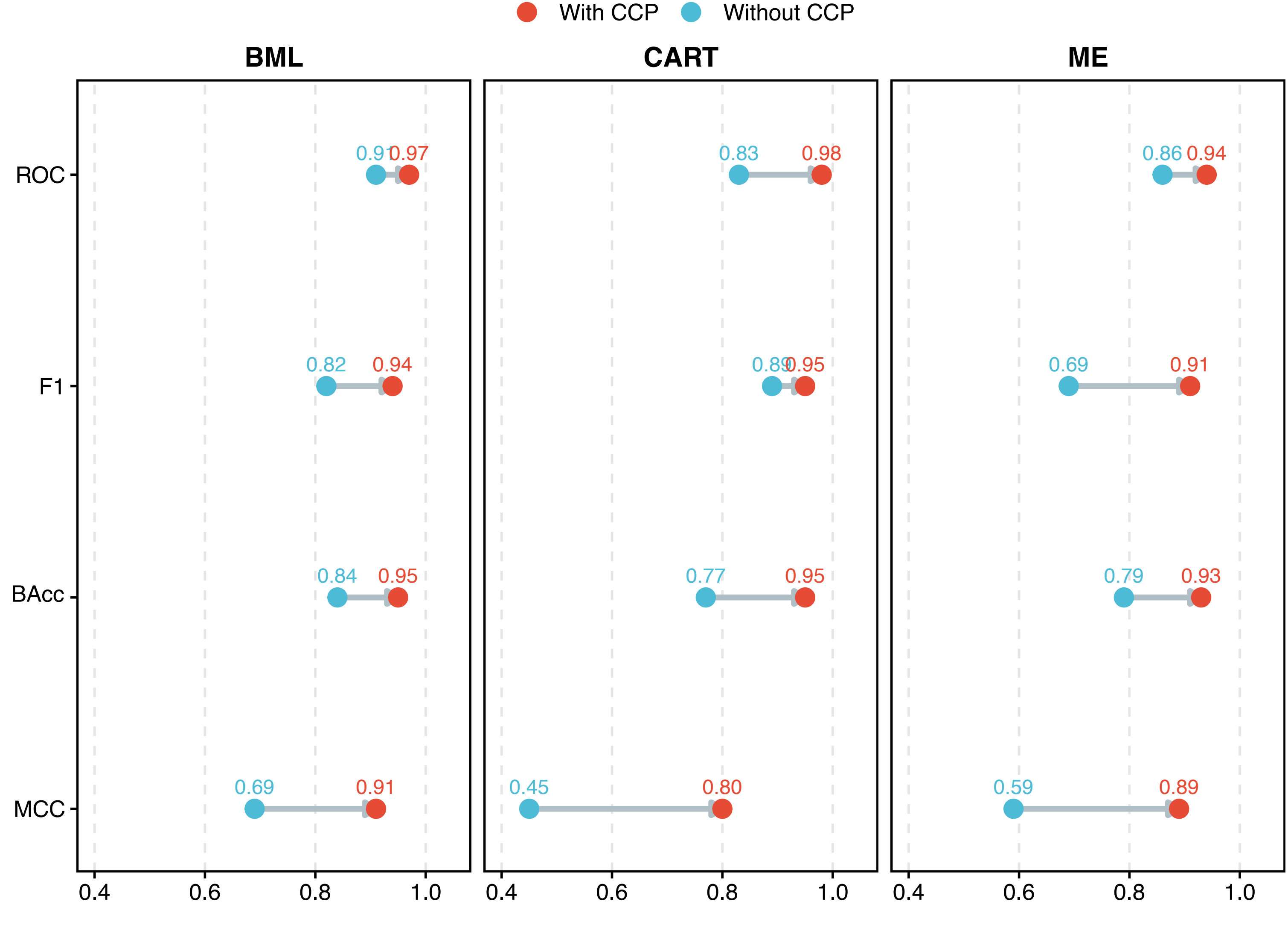
With CCP
Without CCP
BML
CART
ME
ROC
F1
BAcc
MCC
0.91
0.97
0.82
0.94
0.84
0.95
0.69
0.91
0.83
0.98
0.89
0.95
0.77
0.95
0.45
0.80
0.86
0.94
0.69
0.91
0.79
0.93
0.59
0.89
0.4
0.6
0.8
1.0
Performance Score

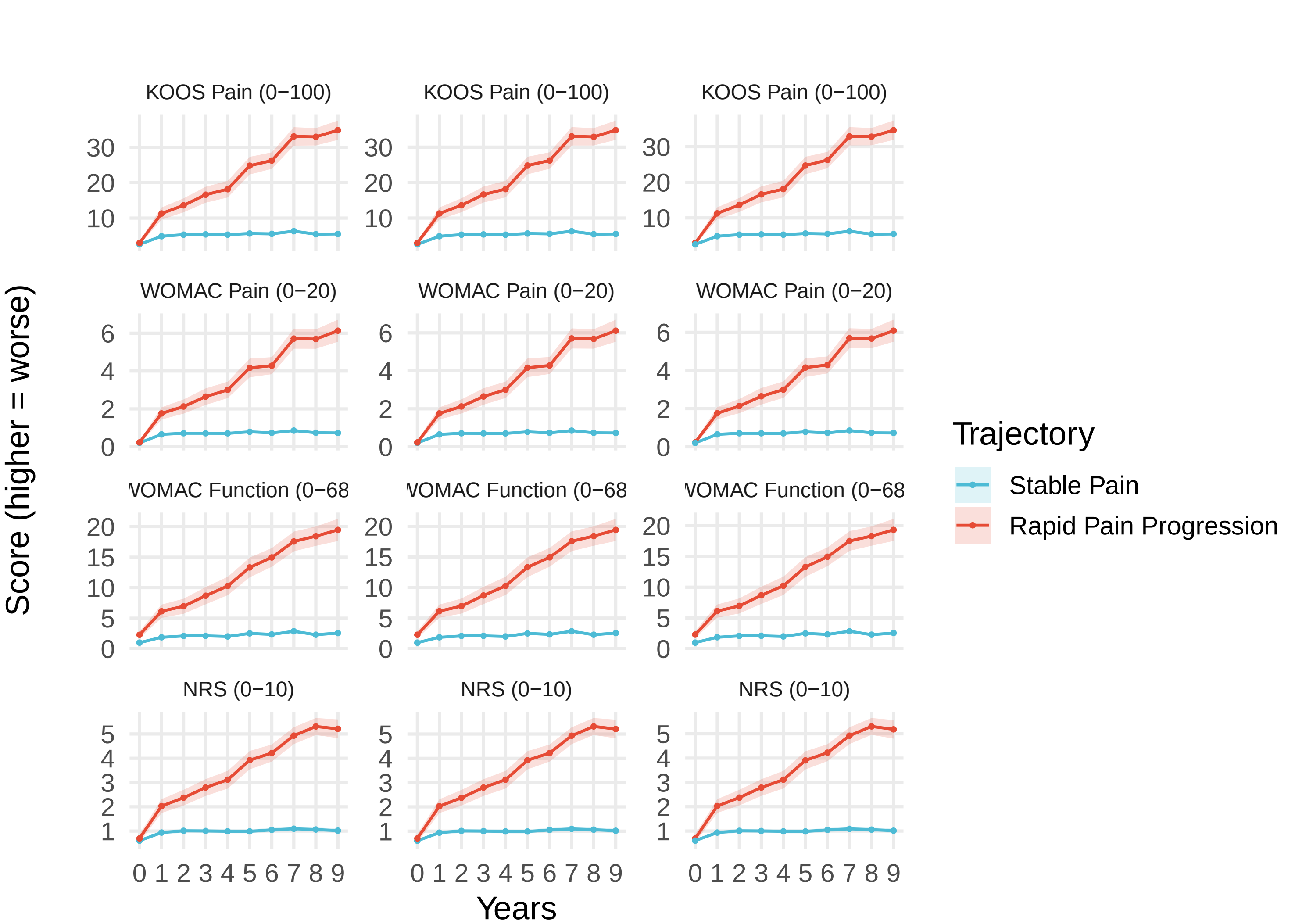

BML model
CART model
ME model
KOOS Pain (0−100)
WOMAC Pain (0−20)
WOMAC Function (0−68
NRS (0−10)
Score (higher = worse)
Years
Trajectory
Stable Pain
Rapid Pain Progression